# Decision Making with Partially Consonant Belief Functions

**Phan H. Giang**
Siemens Medical Solutions
51 Valley Stream Pkwy
Malvern, PA 19355 USA
*phan.giang@siemens.com*

**Prakash P. Shenoy**
University of Kansas School of Business
1300 Sunnyside Ave
Lawrence, KS 66045-7585, USA
*pshenoy@ku.edu*

## Abstract

This paper studies decision making for Walley's partially consonant belief functions (pcb). In a pcb, the set of foci are partitioned. Within each partition, foci are nested. The pcb class includes probability and possibility functions as extreme cases. We adopt an axiomatic system, similar in spirit to von Neumann and Morgenstern's axioms for preferences leading to the linear utility theory, for a preference relation on pcb lotteries. We prove a representation theorem for this preference relation. Utility for a pcb lottery is a combination of linear utility for probabilistic lottery and binary utility for possibilistic lottery.

## 1 Introduction

In recent years, Dempster-Shafer's (DS) belief function theory [3, 11, 13] has drawn an increasing interest in the artificial intelligence and statistics community. The main appeal of the DS theory is its ability to faithfully express a wider class of uncertainty than is expressible by standard probability. Another advantage, as the belief function proponents argue, is that in the DS theory, uncertainty is closely linked to evidence, which is the objective source of uncertainty.

The statistical inference problem is an important background for belief function theory. Dempster [3] and Shafer [12] have demonstrated how belief function theory generalizes Bayesian statistical inference. This generalization allows prior knowledge as well as conditional models to be described by belief functions rather than by probability functions.

The inclusion of Bayesian inference as a special case also gives rise to an opportunity for checking the validity of belief function theory against a number of fundamental principles on which statistics is founded. Walley [14] studies two functionals $Q$ and $R$ that represent, in terms of belief functions, observational and prior evidences, respectively. He finds (see Theorem 3) that in order to be consistent with Bayes rule, observations must be represented by special belief functions. In this class, the set of foci are partitioned. Within each partition, foci are nested. Such belief functions are called *partially consonant* or pcb for short. The partially consonant class is rich enough to include as special cases probability functions (each singleton is a partition) and possibility functions (there is only one element in the partition).

Initially, *consonant* belief functions were used by Shafer [11] to represent statistical evidence. Later, Shafer [12] renounces the idea on the grounds that the set of consonant belief functions is not closed under Dempster's rule of combination. This property is desirable because from a statistical point of view, a series of independent observations can be viewed as a single (compound) observation. So either Dempster's rule is not suitable for combination of independent evidences or the consonant form is not appropriate for representation of evidence. Shafer gives up the latter and keeps the former. However, Walley [14], facing the same choices, comes to a different conclusion. Arguing that (1) the conditions by which Dempster's rule is consistent with Bayes' rule are too restrictive, and (2) Dempster's rule is not unique in satisfying a number of desirable properties, Walley concludes that Dempster's rule is neither suitable for combining independent observations nor for combining prior belief with observational evidence.

One open problem is the use of belief functions for decision making. The main problem here is that a departure from probability also means the loss of Bayesian decision theory which ranks alternatives by maximum expected utilities (EU). A number of proposals for decision making with belief functions have been proposed in literature. One basic idea is to find a transformation that converts a given belief function into a probability function and then use the probability function for decision making [13]. vNM linear utility can be brought to use for belief functions, which could be viewed as lower probabilities [7]. Another approach is to use techniques developed for more general uncertainty measure, e.g., lower prevision [15], capacity [10, 9], which includes belief functions as a special class. We will discuss these works in more details in Section 5.

In this paper, we will propose a decision theory assuming that uncertainty is represented by a pcb. The paper is structured as follows. The derivation of pcb by Walley is reviewed in the next section. In section 3, after a brief review of the vNM axioms that lead to expected utility (EU) representation of probabilistic lotteries as well as the axioms that lead to binary qualitative utility (QU) representation of possibilistic lotteries, we introduce an axiom system for pcb lotteries and prove a representation theorem. We present one example in section 4. Section 5 discusses related literature. The final section has some concluding remarks.

## 2 Partially Consonant Belief Functions

Walley [14] has derived pcb in the context of the statistical inference problem. The statistical inference problem is a triplet $(X, \Theta, F)$ where set $X$ is the sample space, set $\Theta$ is the parameter space, and $P$ is the set of uncertainty measures on $X$ conditioned on parameter values. A statistical evidence (observation) is a value $x \in X$. The prior knowledge about parameters may or may not exist. The objective is to make some inferences about $\Theta$.

The Bayesian theory assumes that (1) $P = \{P_\theta | \theta \in \Theta\}$ is the set of probability functions on $X$ parameterized by elements of $\Theta$; and (2) prior knowledge exists and is represented by a probability function on $\Theta$. The observational evidence and prior knowledge are then combined by Bayes rule.

The likelihood principle (LP) of statistics states that information contained in an observation $x$ is adequately captured by the likelihood function derived from it. The likelihood of a parameter $\theta$ given an observation $x$ is the probability of observing $x$ if $\theta$ is the true parameter i.e., $lik_x(\theta) = P_\theta(x)$. Moreover, proportional likelihood functions are equivalent (see, for example, [1] for a detailed discussion).

Dempster [3], and later Shafer [11, 12], arguing that the Bayesian assumption on the probability form of models and prior knowledge is not always supported by evidence, suggest a more general representation— belief functions—for the purpose. Suppose that $S$ is the set of possible worlds. A *basic probability assignment* or *bpa* is a function

$$m : 2^S \to [0, 1] \quad (1)$$

such that $m(\emptyset) = 0$ and $\sum_{A \subseteq S} m(A) = 1$. The value $m(A)$ can be interpreted as the probability that a world in $A$ will be the *true* world. From a bpa, a number of other functions can be derived

$$Bel(A) \stackrel{def}{=} \sum_{B \subseteq A} m(B) \quad (2)$$

$$Pl(A) \stackrel{def}{=} \sum_{B \cap A \neq \emptyset} m(B) \quad (3)$$

$$Q(A) \stackrel{def}{=} \sum_{A \subseteq B} m(B) \quad (4)$$

*Bel* is referred to as a *belief* function, *Pl* as a *plausibility* function and *Q* as a *commonality* function. These are different forms of the same belief function since starting from any of these forms $(m, Bel, Pl$ or $Q)$, the other three can be recovered.

The combination of independent belief functions is done via Dempster's rule. Suppose $m_1, m_2$ are two belief functions, their combination is another belief function denoted by $(m_1 \oplus m_2)$ defined as follows:

$$(m_1 \oplus m_2)(A) = k^{-1} \sum_{B_1 \cap B_2 = A} m_1(B_1).m_2(B_2) \quad (5)$$

where $k$ is a normalization constant.

In the special case when $m_2$ represents the observation $B$ $(m_2(B) = 1)$, $m_1 \oplus m_2$ is called the conditional of $m_1$ given $B$ $(m_1(\cdot|B))$. In terms of $Pl$, a conditional belief function assumes a familiar form

$$Pl(A|B) = \frac{Pl(A \cap B)}{Pl(B)}. \quad (6)$$

In the statistical inference method argued by Shafer [12], prior knowledge, models and observation are represented in terms of belief functions. The act of inference is done by combining these belief functions using Dempster's rule.

Since prior probability and conditional probabilities in Bayesian model are also belief functions, Walley asks under which conditions their combination by Dempster rule is consistent with Bayes rule. He chooses to work with commonality form for convenience. Specifically, Walley studies functional **Q** that translates likelihood function $lik_x$ which summarizes the Bayesian model $(X, \Theta, F)$ and an observation $x$ into a commonality function $Q$, and functional $\mathbf{R}$ that translates prior probability into a commonality function $R$. There are a number of desirable properties that $\mathbf{Q}, \mathbf{R}$ (two-place mappings) should satisfy. Some technical assumptions are made. $\Theta$ is finite $|\Theta| = N$. $\mathcal{S}$ is the set of likelihood vectors, $\mathcal{P}$ is the set of prior probability vectors. In the axioms listed below (according to Walley's original order), $\tau, \sigma$ stand for arbitrary likelihood vectors in $\mathcal{S}$, $\rho$ for any prior probability in $\mathcal{P}$. Their components are denoted with subscription. $I_B$ is the characterisitic function of subset $B \subseteq \Theta$.

A1 $\mathbf{Q}(\cdot, \tau)$ is a commonality function over $\Theta$.

A2 $\mathbf{Q}(\cdot, \tau) \oplus \mathbf{Q}(\cdot, \sigma) = \mathbf{Q}(\cdot, \tau\sigma)$ if $\tau\sigma \in \mathcal{S}$

A3 $\mathbf{R}(\cdot, \rho)$ is a commonality function over $\Theta$.

A4 If $\rho_j \tau_j > 0$ for some $j$, $\mathbf{R}(\cdot, \rho) \oplus \mathbf{Q}(\cdot, \tau) = \mathbf{R}(\cdot, \tau \circ \rho)$ where $\circ$ denotes Bayes' rule.

A7 $\mathbf{Q}(\cdot, \tau) = \mathbf{Q}(\cdot, c\tau)$ for $0 < c < 1$.

A8 $\mathbf{Q}(\cdot, 1) \oplus \mathbf{Q}(\cdot, \tau) = \mathbf{Q}(\cdot, \tau)$.

A9 If $\tau \in \mathcal{S}$ and $\tau I_B \in \mathcal{S}$ then $\mathbf{Q}(A, \tau I_B) \propto \mathbf{Q}(A, \tau)$ when $A \subseteq B$ and $\mathbf{Q}(A, \tau I_B) = 0$ otherwise.

A1 and A3 convey the idea that belief functions are used to express evidence (observational and prior). A2 requires that two views on multiple independent observations as a compound evidence and as the combination of individual evidence are equivalent. A4 requires that belief function treatment is consistent with Bayesian treatment when applicable. A7 and A8 require the consistency with LP. A9 requires consistency with Bayesian conditioning. Walley has the following theorem:

**Theorem 1 (Walley 1987 [14])** *Assumptions A1, A3, A4, A7, A8 and A9 and a number of technical conditions hold if and only if there is some $\lambda > 0$ and some partition $\{A_1, A_2, \ldots A_s\}$ of $\Theta$ such that for all $\rho \in \mathcal{P}$ and $\tau \in \mathcal{S}$*

$$R(\{\theta_i\}, \rho) = \rho_i^\lambda / \sum_{j=1}^{N} \rho_j^\lambda \quad (7)$$

$$Q(A, \tau) = k(\tau) \min\{\tau_i^\lambda | \theta_i \in A\} \text{ when } A \in \bigcup_{1 \leq i \leq s} 2^{A_i} \quad (8)$$

$$Q(\emptyset, \tau) = 1 \text{ and } Q(A, \tau) = 0 \text{ otherwise} \quad (9)$$

*where $k(\tau) = (\sum_{j=1}^{s} \max\{\tau_i^\lambda | \theta_i \in A_j\})^{-1}$
In addition, assumption A2 is satisfied only if $s = N$.*

Because the satisfaction of Dempster's rule requires evidence presented as probability $(s = N)$, Walley suggests that Dempster's rule cannot be used to combine independent observations. However, all the axioms above (including A2) are satisfied if Dempster's rule is replaced by another rule $(\otimes)$ with the following definition. If $Q_1(A).Q_2(A) \neq 0$ then

$$Q_1 \otimes Q_2(A) \stackrel{\text{def}}{=} k \min\{Q_1(\{\theta\})Q_2(\{\theta\}) | \theta \in A\} \quad (10)$$

otherwise $Q_1 \otimes Q_2(A) \stackrel{\text{def}}{=} 0$. Where $k$ is a constant selected in such a way that makes $Q_{12}$ a commonality function. It is important to note that in special case when $Q_1$ is a pcb and $Q_2$ is an observation $B$ ($Q_2(\theta) = 1$ if $\theta \in B$ and $Q_2(\theta) = 0$ otherwise), Dempster's rule and the new rule has the same effect ($Q_1 \oplus Q_2 = Q_1 \otimes Q_2$).

Parameter $\lambda$ in eq. 8 called the scale parameter. It is used to manipulate the weight of evidence, if necessary. From now on, we assume that $\lambda = 1$. This means, for example, that there is no discounting of evidence. We also choose to work with the plausibility form of a pcb.

$$Pl(A, \tau) = k(\tau) \sum_{j=1}^{s} \max\{\tau_i | \theta_i \in A_j\} \quad (11)$$

A belief function that satisfies equation 8 of theorem 1 is called *partially consonant*. In terms of bpa, it can be seen that the foci are subsets of $A_i$ $1 \leq i \leq s$. Moreover, foci that are subsets of an $A_i$ are nested.

The significance of Walley's result is that (1) it points out the incompatibility between Dempster's rule and the likelihood principle, and (2) isolates a subclass of belief functions and an operation which make a meaningful generalization of Bayesian inference while preserving LP. The pcb class includes probability functions and possibility functions as extreme cases corresponding to $s = N$ and $s = 1$. In the intermediate case when $1 < s < N$, pcb has a nice interpretation. It could be viewed as a model for a number of possibilistic variables conditioned on a probabilistic variable.

Pcb has not received the attention it deserves partly because Walley himself seems to dismiss its usefulness on the ground that the sure-loss or "Dutch book" argument can still be made against the use of pcb in decision making. To reach this conclusion, Walley makes an assumption that associated functions *Bel* and *Pl* of a pcb are interpreted as lower and upper betting rates. In the next section, we will develop a method of making decision with pcb and argue that Walley's assumption is not justified.

## 3 Mixed Utility

In this section, we assume a pcb in the form of a plausibility function $Pl$ on a given partition $\{A_1, A_2, \ldots A_s\}$ of $\Theta$. $A_i = \{\theta_{i1}, \theta_{i1}, \ldots \theta_{in_i}\}$.

$$Pl(A) = \sum_{i=1}^{s} \max\{Pl(\{\theta\}) \mid \theta \in A_i \cap A\} \quad (12)$$

On the algebra $\mathcal{A}$ formed from $\{A_1, A_2, \ldots A_s\}$, $Pl$ is an (additive) probability measure (denoted by $P$) i.e., for $B_1, B_2 \in \mathcal{A}$ and $B_1 \cap B_2 = \emptyset$, $P(B_1 \cup B_2) = P(B_1) + P(B_2)$. Since $\sum_{j=1}^{s} P(A_i) = Pl(\Theta) = 1$, the normalization condition is also satisfied.

Given $A_i$, the conditional plausibility $Pl(\cdot|A_i)$ on the algebra $\mathcal{A}_i$ $1 \leq i \leq s$ formed from the elements of $A_i$, is a possibility measure (denoted by $\Pi_i$) i.e., for $C_1, C_2 \in \mathcal{A}_i$ $\Pi_i(C_1 \cup C_2) = \max\{\Pi_i(C_1), \Pi_i(C_2)\}$, and $\Pi_i(A_i) = Pl(A_i|A_i) = 1$. Thus, we have the following simple theorem. The proof is straightforward.

**Theorem 2** *Plausibility function $Pl$ in eq. 12 induces a probability function $P$ on $\mathcal{A}$ and $s$ (conditional) possibility functions $\Pi_i$ on $\mathcal{A}_i$ $1 \leq i \leq s$. Conversely, given a probability function $P$ on $\mathcal{A}$ and $s$ possibility functions $\Pi_i$ on $\mathcal{A}_i$, the original $Pl$ can be recovered.*

A (probabilistic) act is a mapping $\{A_1, A_2, \ldots A_s\} \to Z$ where $Z$ is the set of prizes. Act $d$ delivers prize $d(A_i)$ in case $A_i$ occurs. An act induces a probabilistic lottery $L$, which is a probability function on $Z$. $L(z_i)$ is the probability of getting prize $z_i$. It is also denoted by $[p_1/z_1, p_2/z_2, \ldots]$ where $p_i = L(z_i)$ and $\sum_i p_i = 1$. Denote by $\mathbf{L}_P$ the set of probabilistic lotteries. The von Neumann and Morgenstern's utility theory (the exposition by Luce and Raiffa [8]) considers a preference relation $\succeq_P$ on $\mathbf{L}_P$ that is assumed to satisfy the following axioms [1]

P1 (Ordering of prizes) Preference $\succeq_P$ on $Z$ is complete and transitive. There are $\overline{z}$ (the best) and $\underline{z}$ (the worst prize) such that $\overline{z} \succ_P \underline{z}$ and for all $z \in Z$, $\overline{z} \succeq_P z$ and $z \succeq_P \underline{z}$.

P2 (Reduction of compound lotteries) Any compound lottery (whose prizes are again lotteries) is indifferent to a simple lottery with prize in $Z$, the probabilities are calculated according to probability calculus.

P3 (Continuity) Each prize $z \in Z$ is indifferent to a canonical lottery involving just $\overline{z}$ and $\underline{z}$.

P4 (Substitutability) In any lottery, each prize can be replaced by the canonical lottery that is indifferent to it.

P5 (Transitivity) $\succeq_P$ on $\mathbf{L}_P$ is transitive.

P6 (Monotonicity) $[p/\overline{z}, (1-p)/\underline{z}] \succeq_P [p'/\overline{z}, (1-p')/\underline{z}]$ iff $p \geq p'$.

**Theorem 3 (von Neumann & Morgenstern)**
*$\succeq_P$ satisfies axioms P1 – P6 iff there exists a utility function $u : \mathbf{L}_P \to [0,1]$ such that $L_1 \succeq L_2$ iff $u(L_1) \geq u(L_2)$. Moreover, $u$ is unique up to a positive affine transformation and it has the form*

$$u([p_1/z_1, p_2/z_2, \ldots, p_n/z_n]) = \sum_{i=1}^{n} p_i \cdot u(z_i).$$

Following vNM approach, in [6] we develop a utility theory for possibilistic lotteries. Suppose $W = \{w_1, w_2, \ldots w_k\}$ is the set of prizes including the best ($\overline{w}$) and the worst ($\underline{w}$) prizes. A possibilistic lottery is a possibility function on $W$. It is also denoted by $[\pi_1/w_1, \pi_2/w_2, \ldots]$ where $\pi_i$ is the possibility of getting prize $w_i$. $\max_i \pi_i = 1$. The set of possibilistic lotteries is denoted by $\mathbf{L}_\Pi$. We require that a preference relation $\succeq_\Pi$ on $\mathbf{L}_\Pi$ satisfies the following axioms.

Π1 (Ordering of prizes) Preference $\succeq_\Pi$ on $W$ is complete and transitive. There are $\overline{w}$ (the best) and $\underline{w}$ (the worst prize) such that $\overline{w} \succ_\Pi \underline{w}$ and for all $w \in W$, $\overline{w} \succeq_\Pi w$ and $w \succeq_\Pi \underline{w}$.

Π2 (Reduction of compound lotteries) Any compound lottery (whose prizes are again lotteries) is indifferent to a simple lottery with prize in $W$, the possibilities are calculated according to possibility calculus.

Π3 (Continuity) Each prize $w \in W$ is indifferent to a canonical possibilistic lottery involving just $\overline{w}$ and $\underline{w}$.

Π4 (Substitutability) In any lottery, each prize can be replaced by the canonical lottery that is indifferent to it.

Π5 (Transitivity) $\succeq_\Pi$ on $\mathbf{L}_\Pi$ is transitive.

Π6 (Monotonicity) $[\lambda/\overline{w}, \rho/\underline{w}] \succeq_\Pi [\lambda'/\overline{w}, \rho'/\underline{w}]$ iff $\lambda \geq \lambda'$ and $\rho \leq \rho'$.

**Theorem 4 (Giang & Shenoy 2002)** *$\succeq_\Pi$ satisfies axioms $\Pi_1 - \Pi_6$ iff there exists a utility function $qu : \mathbf{L}_\Pi \to \mathcal{B}$ such that for $\Pi_1, \Pi_2 \in \mathbf{L}_\Pi$, $\Pi_1 \succeq_\Pi \Pi_2$ iff $qu(\Pi_1) \geqslant qu(Pi_2)$. Moreover, qu has the form*

$$qu([\pi_1/w_1, \pi_2/w_2, \ldots, \pi_m/w_m]) = \max_{1 \leq i \leq m}\{\pi_i.qu(w_i)\}$$

---
[1] Derivative forms of strict preference ($\succ_P$) and indifference ($\sim_P$) are defined as usual.

where $\mathcal{B} \stackrel{def}{=} \{<\lambda,\rho> | \lambda, \rho \in [0,1], \max(\lambda,\rho) = 1\}$;
for $<\lambda,\rho>, <\lambda',\rho'> \in \mathcal{B}$, order $\geqslant$ and max are defined as follows:
$<\lambda,\rho> \geqslant <\lambda',\rho'>$ iff $\lambda \geq \lambda'$ and $\rho \leq \rho'$;
$\max\{<\lambda,\rho>, <\lambda',\rho'>\} \stackrel{def}{=} <\max(\lambda,\lambda'), \max(\rho,\rho')>$;

The difference between this theorem and one by vNM is that for possibilistic lotteries the binary utility scale is used. The constrast between the binary utility for a possibilistic lottery and the real utility for a probabilistic lottery has a parallel in the different formats that uncertainty is expressed. In possibility theory, uncertainty of an event is a pair of necessity and possibility. However, the probability of an event is a single real number. Interested readers are referred to [5] for a more detailed discussion. The binary utility scale is a set of pairs of number in the unit interval so that the maximum of two numbers is 1. The order $\geqslant$ between pairs is such that a pair is better than the other if the left component (a real number) of the former is larger than the left component of the later and the right component of the former is smaller than corresponding number of the latter. So the top element of $\mathcal{B}$ is $<1, 0>$ and the bottom element is $<0, 1>$.

The operation max applied to pairs will produce a new pair whose components are component-wise maxima. It is not used to choose the better (according to $\geqslant$) among pairs.

In our approach, the possibility degree of an event - plausibility of a consonant belief function - is not treated as an upper betting rate as suggested by Walley [15]. In [6, 5], we propose a framework (likelihood gambles) in which a person bets given a possibilitic lottery. A unknown parameter $\theta$ has two possible values $\{\theta_1, \theta_2\}$. For each point in the sample space $x \in X$, one has (normalized) likelihoods $Lik_x(\theta_1)$ and $Lik_x(\theta_2)$. Unlike a standard gamble in which rewards are pegged with the observation, in a likelihood gamble the rewards are pegged with the values of the unknown parameter. Specifically, given an observation $x$, what is a price that a person would be willing to pay for a contract that gives her \$1 if the true value of $\theta$ is $\theta_1$ and nothing otherwise. Obviously, Walley's sure-loss argument against the use of pcb is void with respect to likelihood gambles.

From the behavior of a decision maker toward likelihood gambles, one can deduce her *implicit prior probabilities* (of the event that true value of the parameter is $\theta_1$). The idea is that given such priors, the decision maker will calculate the posterior probability on the parameter space. And thus, she converts the likelihood gamble to a standard gamble. In our framework, the implicit priors are not required to remain invariant in different intances of likelihood gambles (corresponding to different sample points). But instead, it can vary as long as monotonicity axiom (II6) is satisfied. For more details, readers are referred to [5]. Implicit priors can be used to differentiate decision maker's *attitudes toward ambiguity* (uncertainty) in contrast to the *attitudes toward risk* which are represented by the shape of utility functions[2]. An implicit prior of less than .5 means *ambiguity aversion*. (It means that one is willing to pay less than \$.50 for a "fair" likelihood gamble $[1/\$1, 1/\$0]$ in which the likelihood of $\theta_1$ is equal the likelihood of $\theta_2$.) A value larger than .5 is *ambiguity seeking* and .5 means *ambiguity neutral*.

Let us now return to our problem. We want to make a decision in a uncertain situation described by pcb $Pl$. An act is a mapping $d : \Theta \to W$ where $W$ is the set of prizes. We also summarize act $d$ in the form of a lottery $[a_1/w_1, a_2/w_2, \ldots, a_k/w_k]$ where $a_i = Pl(d^{-1}(w_i))$ i.e., $a_i$ is the plausibility of an event whose occurences associated by act $d$ with prize $w_i$. Notice that unlike probabilistic or possibilistic lotteries, there is no explicit normalization condition for $a_i$ $1 \leq i \leq k$. Denote by $\mathbf{L}_B$ the set of lotteries. We consider preference relation $\succeq$ on $\mathbf{L}_B$ and list a number of desirable properties along the line of vNM axioms.

$B1$ (Ordering of prizes) Preference $\succeq$ on $W$ is complete and transitive. There are $\overline{w}$ (the best) and $\underline{w}$ (the worst prize) such that $\overline{w} \succ \underline{w}$ and for all $w \in W$, $\overline{w} \succeq w$ and $w \succeq \underline{w}$.

$B2$ (Reduction of compound lotteries) Any compound lottery (whose prizes are again lotteries) is indifferent to a simple lottery with prize in $W$, the plausibilities are calculated according to belief function calculus.

$B3$ (Continuity of prize) Each prize $w \in W$ is indifferent to a canonical possibilistic lottery involving $\overline{w}$ and $\underline{w}$.

$B4$ (Substitutability) Indifferent lotteries can replace each other in any lottery.

$B5$ (Transitivity) $\succeq$ on $\mathbf{L}_B$ is transitive.

$B6$ (Monotonicity) For canonical lotteries $L_1 = [\lambda/\overline{w}, \rho/\underline{w}]$ and $L_2 = [\lambda'/\overline{w}, \rho'/\underline{w}]$, $L_1 \succeq L_2$ iff $\lambda \geq \lambda'$ and $\rho \leq \rho'$.

$B7$ (Equivalence between forms) Each canonical possibilistic lottery is indifferent to a canonical probabilistic lottery.

$B1$, the same as $P1$ or $I1$, is about order on the set of prizes. $B2$ means the following. Let us consider a

---

[2]Convexity means risk seeking, concavity means risk aversion and linearity means risk neutral.

compound lottery $L = [a_1/L_1, a_2/L_2, \ldots a_k/L_k]$ where $a_i = Pl(E_i)$ i.e., the pcb plausibility of an event $E_i$ associated with lottery-reward $L_i$, $1 \leq i \leq k$. Further suppose $L_i = [b_{i1}/w_1, b_{i2}/w_2, \ldots b_{in}/w_n]$ where $b_{ij} = Pl(E_{ij}|E_i)$ i.e., conditional pcb plausibility of an event $E_{ij}$ associated with prize $w_j$. (Here $E_{ij}$ is defined as the set of states that $L_i$ will deliver $w_j$. $E_{ij}$ can be empty.) Thus, if event $S_j \stackrel{def}{=} \cup_{i=1}^{k}(E_i \cap E_{ij})$ occurs then the lottery holder is rewarded with prize $w_j$. Therefore, lottery $[c_1/w_1, c_2/w_2 \ldots c_k/w_k]$ where $c_j = Pl(S_j)$ should be considered indifferent to the original compound $L$. Axiom $B2$ requires just that.

If $Pl$ is a *consonant* belief function on $\Theta$, a lottery induced from an act, which delivers either $\overline{w}$ or $\underline{w}$ for each $\theta \in \Theta$ is called a *canonical possibilistic* lottery. Axiom $B3$ requires that each prize $w \in W$ is indifferent to one of such lotteries.

If $Pl$ is a probability function on $\Theta$, an act that delivers either $\overline{w}$ or $\underline{w}$ for each $\theta \in \Theta$ induces a lottery called a *canonical probabilistic*. $B7$ requires that each canonical possibilistic lottery is indifferent to a canonical probabilistic lottery. In other words, the decision maker must be able to switch between uncertainty forms attached to the prizes. This condition is strong but not unreasonable. For example, the most preferred canonical probabilistic lottery $[1/\overline{w}, 0/\underline{w}]$ is naturally considered as good as the best prize $\overline{w}$, which, in turn, is considered the same as the most preferred canonical possibilistic lottery $[1/\overline{w}, 0/\underline{w}]$. The argument works also for the worst prize i.e., we can reasonable claim $[0/\overline{w}, 1/\underline{w}] \sim \underline{w}$ no matter how 1 and 0 are interpreted (as probability or as possibility). Thus, the range covered by canonical probabilistic lotteries is the same as one covered by canonical possibilistic lotteries.

It is should be noted that monotonicity axiom $B6$, as the order requirement, holds for both types of probabilistic and possibilistic comparison i.e., the numbers can be interpreted either as probability or possibility.

**Theorem 5** *If $\succeq$ on $L_B$ satisfies axioms $B1 - B7$, then there exist functions $qu: L_\Pi \to \mathcal{B}$, $u: L_B \to [0,1]$ and $t: \mathcal{B} \to [0,1]$ such that $L_1 \succeq L_2$ iff $u(L_1) \geq u(L_2)$. Moreover, for $L = [p_1/L_1, p_2/L_2, \ldots p_s/L_s]$ and $L_i = [\pi_{i1}/w_1, \pi_{i2}/w_2, \ldots \pi_{ik}/w_k]$ where $p_i$ are probabilities and $\pi_{ij}$ are possibilities*

$$u(L) = \sum_{i=1}^{s} p_i . t(\max_{1 \leq j \leq k} \{\pi_{ij}.qu(w_j)\}) \quad (13)$$

**Proof:** Suppose lottery $L$ is induced by act $d$ and pcb $Pl$ given in eq. 12. By Theorem 2, $Pl$ is decomposed into a set of a probability function $P$ and $s$ conditional possibility functions $\Pi_i$ $1 \leq i \leq s$.

By axiom $B2$, $L \sim L'$ where $L'$ is the compound lottery $[p_1/L_1, p_2/L_2, \ldots p_s/L_s]$, where $p_i = P(A_i)$ for $1 \leq i \leq s$. $L_i$ are possibilistic lotteries $[\pi_{i1}/w_1, \pi_{i2}/w_2, \ldots \pi_{ik}/w_k]$ where $\pi_{ij} = \Pi_i(d^{-1}(w_j) \cap A_i)$.

Notice that $L_\Pi \subseteq L_B$, we show $\succeq_\Pi$, which is $\succeq$ restricted to $L_\Pi$, satisfies axioms $\Pi 1 - \Pi 6$ if $\succeq$ satisfies $B1 - B7$. It is easy to note that $B1$ and $B3$ are $\Pi 1$ and $\Pi 3$ respectively. When $Pl$ is consonant, $B2$ reduces to $\Pi 2$. $B4$ is slightly more general and therefore implies $\Pi 4$. Since $\succeq$ on $L_B$ is transitive, $\succeq_\Pi$ on $L_\Pi$ is transitive too. Thus, $\Pi 5$ is satisfied. $B6$ reduces to $\Pi 6$ for possibilistic lotteries. By Theorem 4, there is a binary utility function $qu: L_\Pi \to \mathcal{B}$ representing $\succeq_\Pi$ and for $L_i$

$$qu(L_i) = \max_{1 \leq j \leq k} \{\pi_{ij}.qu(w_j)\} \quad (14)$$

Suppose $qu(L_i) = <\lambda_i, \rho_i>$. Then it is not difficult to show $L_i \sim_\Pi [\lambda_i/\overline{w}, \rho_i/\underline{w}]$. By axiom $B7$ $[\lambda_i/\overline{w}, \rho_i/\underline{w}] \sim [q_i/\overline{w}, (1-q_i)/\underline{w}]$ for some $0 \leq q_i \leq 1$. By transitivity of $\succeq$ ($B5$), $L_i \sim [q_i/\overline{w}, (1-q_i)/\underline{w}]$. By substitutability ($B4$), $[q_i/\overline{w}, (1-q_i)/\underline{w}]$ can substitute $L_i$ in $L'$ to obtain an indifferent $L''$, which is a probabilistic lottery. By reduction of compound lottery $B2$, $L''$ is reduced to a canonical probabilistic lottery $[q/\overline{w}, (1-q)/\underline{w}]$. Let us denote the set of probabilistic lotteries constructed with the set of prize $Z$ equal to $\{\overline{w}, \underline{w}\}$ by $L_P^2$. $\succeq$ restricted on $L_P^2$ is denoted by $\succeq_P$.

It is not difficult to verify that axioms $P1 - P6$ are satisfied by $\succeq_P$. $P1$ is obvious because $Z$ has two elements $\overline{w}$ and $\underline{w}$. $B2$ restricted for probabilistic lotteries reduces to $P2$. $P3$ is trivially satisfied because $Z$ has only two elements. $P4$ is satisfied due to $B4$. $\succeq$ is transitive, so $\succeq_P$ is too. Thus, $P5$ is satisfied. $B6$ applied to probabilistic lotteries implies $P6$. By Theorem 3, $\succeq_P$ is represented by a utility function

$$u([q/\overline{w}, (1-q)/\underline{w}]) = q.u(\overline{w}) + (1-q).u(\underline{w}) \quad (15)$$

By transitivity, this is also the utility of the original pcb lottery. Let us make a convenient assumption that $u(\overline{w}) = 1$ and $u(\underline{w}) = 0$ and formalize equivalence relationship between canonical possibilistic and canonical probabilistic lotteries by a function $t: \mathcal{B} \to [0,1]$ defined by $t(<\lambda, \rho>) = p$ iff $[\lambda/\overline{w}, \rho/\underline{w}] \sim [p/\overline{w}, (1-p)/\underline{w}]$. We can rewrite the equation 15 for lottery $L = [p_1/L_1, p_2/L_2, \ldots p_s/L_s]$ and $L_i = [\pi_{i1}/w_1, \pi_{i2}/w_2, \ldots \pi_{ik}/w_k]$ as

$$u(L) = \sum_{i=1}^{s} p_i . t(\max_{1 \leq j \leq k} \{\pi_{ij}.qu(w_j)\}) \quad (16)$$

This completes the proof. ∎

A prize $w$ can be viewed as a lottery induced by act $d_w(\theta) = w$ for a $\theta \in \Theta$ and a pcb $m(\{\theta\}) = 1$. It means there is only one partition, $s = 1$, and $p_1 = 1$. There is also one $\pi_{ij} = 1$ (corresponding to $\theta$), and $\pi_{ij} = 0$ for other. Applying eq. 13 for that lottery, we have

$$u(w) = t(qu(w)) \qquad (17)$$

With this equality, formula in eq. 13 reduces to vNM linear utility for probabilistic lotteries

$$u([p_1/w_1, p_2/w_2, \ldots p_k/w_k]) = \sum_{i=1}^{k} p_i \cdot u(w_i)$$

By definition of function $t$, $(t(<\lambda,\rho>) = p$ iff $[\lambda/\overline{w}, \rho/\underline{w}] \sim [p/\overline{w}, (1-p)/\underline{w}])$, and axiom $B6$, it is clear that $t$ is a strictly increasing function i.e., for $b_1, b_2 \in \mathcal{B}$, $t(b_1) \geq t(b_2)$ iff $b_1 \gtrapprox b_2$. For possibilistic lotteries, formula in eq. 13 reduces to

$$u([\pi_1/w_1, \pi_2/w_2 \ldots \pi_k/w_k]) = t(\max_{1 \leq j \leq k} \{\pi_j \cdot qu(w_j)\})$$

Thus, the ranking of lotteries by $u$ is the same as ranking by binary utility in Theorem 4.

## 4 An Example

In this example, we will offer our treatment of Ellsberg's paradox [4]. In an urn, there are 90 balls of the same size. The balls are painted one of three colors: red, yellow and white. It is known that exact 30 balls are red. The proportions of yellow and white are not known.

We consider four gambles. Gamble $IA$ offers \$1 if a randomly drawn ball is red, nothing otherwise. Gamble $IB$ offers \$1 if the ball is yellow, nothing otherwise. Gamble $IIA$ offers \$1 if a randomly drawn ball is red or white, nothing if the ball is yellow. Gamble $IIB$ offers \$1 if the ball is yellow or white and nothing if it is red.

The uncertainty in the problem is nicely described by the following pcb. $m(\{red\}) = \frac{1}{3}$ and $m(\{yellow, white\}) = \frac{2}{3}$. This will decompose into a probability function $P(\{red\}) = \frac{1}{3}$ and $P(\{yellow, white\}) = \frac{2}{3}$ and a conditional possibility function $\Pi(yellow|\{yellow, white\}) = \Pi(white|\{yellow, white\}) = 1$.

We assume binary utility function $qu(\$1) = <1,0>$, $qu(\$0) = <0,1>$ and $qu(.4) = <1,1>$. The first two equalities are natural since \$1 is the best outcome and \$0 is the worst outcome. The last one also means that the implicit prior probability of getting \$1 in a likelihood test is .4. This value, being less than .5, indicates a slight uncertainty aversion.

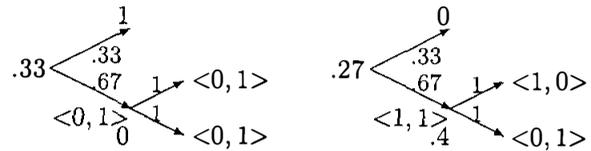

Figure 1: Ellsberg's lotteries IA and IB

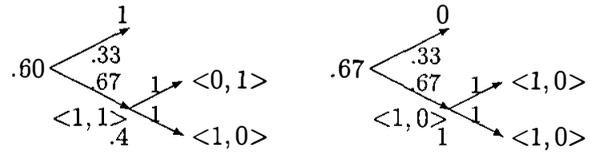

Figure 2: Ellsberg's lotteries IIA and IIB

Ellsberg found that a sizable proportion of respondents prefered $IA$ to $IB$ and, at the same time, prefered $IIB$ to $IIA$. This is a paradox because the observed preference violates Savage's sure-thing principle.

In Figures 1 and 2 we show the calculation of mixed utility for the gambles. We have $u(IA) = .33$, $u(IB) = .27$, $u(IIA) = .60$ and $u(IIB) = .67$. This is consistent with the observed behavior.

## 5 Related Works

Smets [13] argues for a two-level process. At the *credal* level, an agent uses belief functions to represent and to reason with uncertainty. When there is a need to make a decision, the agent moves to another *pignistic* level in which the belief function is tranformed to a probability function. The vNM expected utilities for alternatives are calculated with respect to this probability function. Specifically, in the pignistic transformation, the mass that assigned to a subset is divided equally to each element in the set. For instance, the pignisitic transformation of the pcb in Ellsberg's paradox is a probability function

| $\Theta$ | red | yellow | white |
|---|---|---|---|
| $P_{Bet}$ | .33 | .33 | .33 |

With respect to $P_{Bet}$ the utilities of the four lotteries are $u(IA) = u(IB) = .33$ and $u(IIA) = u(IIB) = .66$. This does not explain Ellsberg's paradox. It should be noted however, that this preference would be observed in our framework for a uncertainty-neutral decision maker who equated a "fair" likelihood lottery $[1/\$1, 1/\$0]$ with \$.50.

Although Smets has argued otherwise, the pignistic probability function is obtained by applying the principle of insufficient reasoning for non-singleton foci. Ironically, the main motivation for using belief functions is the desire to avoid this ad hoc principle.

A discussion of pignistic transformation can be found in Cobb and Shenoy [2]. They argue for the use of plausibilistic transformation in which a probability function is obtained by normalization of plausibility values of singletons. For instance, plausibilistic transformation for the same pcb Ellsberg's example is

| $\Theta$ | red | yellow | white |
|---|---|---|---|
| $P_{Pla}$ | .20 | .40 | .40 |

With respect to $P_{Pla}$, the utilities of the four lotteries are $u(IA) = .20$, $u(IB) = .40$ and $u(IIA) = .60$, $u(IIB) = .80$ These utilities are not able to explain Ellsberg's paradox. Also, it suggests that $IB \succ IA$, which is contrary to the observed empirical behavior. The plausibility transformation has a number of drawbacks. Notice that the probabilities assigned to singletons in the original pcb are modified downward after translation. The magnitude of distortion depends on (1) the total of masses assigned to non-singletons and (2) the sizes of non-singleton foci. However, it is possible to argue that the plausibility transformation is for the Dempster-Shafer theory of belief functions, in which it is inappropriate to interpret belief and plausibility functions as lower and upper bounds on some true but unknown probabilities.

Walley [15] studies a class of imprecise probabilities: lower prevision and, its dual, upper prevision. This class includes belief functions as a subclass. He argues that imprecise probability allows only a partial preference ordering among alternatives. The most one can make from such a partial order is to exclude all dominated alternatives. The set of remaining alternatives, which may be large, is left to decision maker to choose by calling in an additional choice mechanism e.g., randomization. This indeterminacy is a significant inconvenience to the decision maker. This approach, developed for imprecise probability in general, also fails to take into account the specific structure offered by pcb.

Jaffray [7], using lower probability semantics for belief function, shows that the axioms of vNM theory augmented with a number of assumptions leads to Hurwicz's $\alpha$−criteria. Utility of a belief function lottery is a weighted average of lower and upper expected utilities

$$u_{Bel}(d) = \alpha . \inf_{P \in \mathbf{P}} u_P(d) + (1-\alpha) . \sup_{P \in \mathbf{P}} u_P(d) \quad (18)$$

where $\mathbf{P}$ is the set of probability functions whose lower envelope is the belief function. $0 \leq \alpha \leq 1$ is interpreted as a pessimism index. Some drawbacks of this approach is that (1) it is expensive to work with a set (may be large) of probability functions; and (2) little is known about how to pick $\alpha$.

Schmeidler [10], Sarin & Wakker [9] argue for the use of Choquet expected utility (CEU) for *non-additive probability* or capacity. A real value set function $v$ on $\Theta$ is a *capacity* if it satisfies normalization conditions ($v(\emptyset) = 0$, $v(\Theta) = 1$) and monotonicity ($v(A) \leq v(B)$ if $A \subset B$). Obviously, capacity includes lower prevision studied by Walley [15], which in turn includes belief functions as a special class.

For simplicity, we assume the prizes in $W$ are measured in utility (in the unit interval) and are ordered $w_1 > w_2 > \ldots > w_k$. Assume a dummy $w_{k+1} = 0$. For a decision $d$ with $d^{-1}(w_i) = E_i$, $1 \leq i \leq k$, Choquet expected utility given capacity $v$ is defined as

$$CEU(d) = \sum_{i=1}^{k} (w_i - w_{i+1}).v(\cup_{j=1}^{i} E_j) \quad (19)$$

The CEU representation is obtained by relaxing a number of axioms that originally leads to vNM expected utility representation for probabilistic lotteries. The appeal of CEU is that it leads to an (complete) order of alternatives and is supported by intuitive axioms. However, when applied to belief function, use of CEU is not satisfactory on two accounts. First, like Walley's proposal, it fails to take into account the additional structure that belief functions, but not non-additive probability in general, possesses. A more serious shortcoming that makes CEU unsuitable for belief functions is the fact that the ranking of lotteries by CEU depends on the forms in which uncertainty is represented. A belief function exists in many equivalent forms: bpa ($m$), belief ($Bel$), plausibility ($Pl$) and commonality ($Q$). It is easy to check by definitions of $Bel$ (eq. 2) and $Pl$ (eq. 3) that both satisfy the definition of a non-additive probability. Obviously, the ranking by CEU with respect to $Bel$ is different from one by CEU with respect to $Pl$. In contrast, in our approach, $qu$ avoids this shortcoming by using binary utilities.

Let us calculate *CEU* with respect to *Bel* and *Pl* for the lotteries in Ellsberg's example. For $IA$, $E_1 = \{red\}$, $E_2 = \{yellow, white\}$ because red is associated with $1 prize and yellow, white with zero. Using eq. 19, we find $CEU_{Bel}(IA) = CEU_{Pl}(IA) = .33$. For $IB$, we have $E_1 = \{yellow\}$, $E_2 = \{red, white\}$. We find $CEU_{Bel}(IB) = 0$ and $CEU_{Pl}(IB) = .67$. Therefore, $CEU_{Bel}$ ranks $IA \succ IB$. However, $CEU_{Pl}$ ranks $IB \succ IA$.

## 6 Summary and Conclusions

In this paper we study decision making with a special class of belief functions called partially consonant belief functions or pcb, which was introduced by Walley [14]. Pcb is important because (1) it offers a meaningful generalization of both probability and possibility (those are extreme cases); (2) it is the only subclass of belief functions, which is consistent with a number of fundamental principles of statistics. Pcb has a nice interpretation– it can be decomposed into a probability function and a number of conditional possibility functions.

We use an axiomatic approach for the problem. Our axiomatics is similar to and inspired by von Neumann - Morgenstern's linear utility theory. We prove a representation theorem for a preference relation on pcb lotteries. Utility for a pcb lottery is a mixed construct of linear utility and binary utility.

## Acknowledgments

This paper was written when Phan Giang was a doctoral student at the University of Kansas and was supported by a Research Assistantship from the School of Business. The authors would like to thank UAI referees for constructive suggestions to improve this paper.